\pdfoutput=1

\documentclass[10pt]{article} 
\PassOptionsToPackage{dvipsnames}{xcolor}
\usepackage[preprint]{tmlr}

\usepackage{amsmath}
\usepackage{amssymb}
\usepackage{booktabs}
\usepackage{etoolbox}
\usepackage{float}
\usepackage{graphicx}
\usepackage{subcaption}
\usepackage{tabularray}
\usepackage{tabulary}
\usepackage{wrapfig}
\usepackage{xspace}
\usepackage{tikz}
\usetikzlibrary{arrows.meta}
\usetikzlibrary{trees}
\usepackage{forest}
\definecolor{taskdp}{HTML}{56B4E9} 
\definecolor{taskpl}{HTML}{E69F00} 
\definecolor{trace}{HTML}{009E73}  
\definecolor{sp_token}{HTML}{6E6E6E} 
\usepackage{hyperref}
\usepackage{url}
\usepackage{cleveref}
\usepackage[acronym]{glossaries-extra}
\glsdisablehyper
\setabbreviationstyle[acronym]{long-short}
\makeglossaries

\crefname{table}{Tab.}{Tabs.}
\Crefname{table}{Tab.}{Tabs.}
\crefname{figure}{Fig.}{Figs.}
\Crefname{figure}{Fig.}{Figs.}
\crefname{section}{Sec.}{Secs.}
\Crefname{section}{Sec.}{Secs.}
\crefname{equation}{Eq.}{Eqs.}
\Crefname{equation}{Eq.}{Eqs.}

\newcommand{\eg}{e.g.\@\xspace}
\newcommand{\ie}{i.e.\@\xspace}

\newacronym{llm}{LLM}{large language model}
\newacronym[longplural=deterministic finite automata]{dfa}{DFA}{deterministic finite automaton}
\newacronym[shortplural=iid]{iid}{i.i.d.}{independent and identically distributed}
\newacronym{cot}{CoT}{chain of thought}
\newacronym{cfg}{CFG}{context-free grammar}
\newacronym{dp}{DP}{deepest path}
\newacronym{pl}{PL}{maximum-order preleaf}
\newacronym{icl}{ICL}{in-context learning}
\newacronym{sft}{SFT}{supervised fine-tuning}
\newacronym{moe}{MoE}{mixture of experts}
\newacronym{ssm}{SSM}{state-space model}
\newacronym{rope}{RoPE}{rotary positional encoding}

\newcommand{\glsforcefull}[1]{
  \glsentrylong{#1} (\glsentryshort{#1})
}

\NewColumnType{L}{l}
\NewColumnType{C}{c}
\NewColumnType{R}{r}
\NewColumnType{V}{|[wd=0.4pt]}
\NewColumnType{D}{|[wd=0.4pt]|[wd=0.4pt]}

\NewTableCommand\btvheadrule{\hline[wd=0.55pt]}
\NewTableCommand\btvfootrule{\hline[wd=0.55pt]}

\NewDocumentEnvironment{btvtable}{m}
  {\begin{tblr}{
      colspec={#1},
      hline{1,Z} = {wd=1pt},
      rowsep=3pt,
      colsep=6pt,
      cells = {valign=m},
    }}
  {\end{tblr}}

\AtBeginEnvironment{wrapfigure}{\setlength{\intextsep}{1\normalbaselineskip}}

\title{Protoreasoning in Tiny Transformers}

\author{\name Eduardo Valle \addr Fin AI Research \\ \name Fergal Reid \addr Fin AI Research}

\def\tmlrmonth{05}  
\def\tmlryear{2026} 
\def\openreview{\url{https://openreview.net/forum?id=XXXX}} 

\begin{document}

\maketitle

\begin{abstract}
We show that tiny transformers can profitably employ a simple form of \glsxtrlong{cot}, which we call protoreasoning,
allowing us to study step-by-step reasoning on $\sim$1M-parameter models and opening up opportunities for much more detailed experimentation and analysis than is feasible for larger models.
Current \glsxtrlongpl{llm} exhibit impressive step-by-step reasoning, but we have yet to understand its generality, \ie, when and how \glsentryshortpl{llm} learn genuinely general algorithms rather than ``bags of heuristics.'' Such questions are hard to settle on compute-intensive frontier models trained on opaque data.
To work at model scales far below the threshold for natural-language competence, we define reasoning-friendly tasks on Dyck languages (sentences of correctly nested brackets).
We find that protoreasoning traces substantially close the out-of-distribution generalization gap, and ablations confirm that the trace's content, not merely its extra tokens, drives the gain. \end{abstract}

\glsresetall
\glsunset{dp}\glsunset{pl}

\section{Introduction}
\label{sec:introduction}
\Gls{cot} reasoning is crucial for getting the best out of \glspl{llm}, yet hard to study on frontier models too costly to train repeatedly. Small transformers are far more wieldy for comprehensive experimentation and interpretability, but sit orders of magnitude below the scale at which natural-language reasoning becomes viable. We bridge that gap, inducing tiny models ($\sim$1M parameters) to reason in a primitive but useful way.

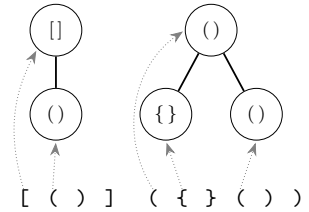
\begin{wrapfigure}{r}{0.25\textwidth}
    \centering
    \resizebox{\linewidth}{!}{
    \begin{tikzpicture}[
        every node/.style={inner sep=0pt},
        bnode/.style={circle, draw, minimum size=8mm, font=\footnotesize},
        edge/.style={thick},
        link/.style={-{Stealth[length=2mm]}, gray, densely dotted, shorten >=1pt},
        arc/.style={gray, semithick, shorten >=1pt, shorten <=1pt},
        chr/.style={font=\ttfamily\normalsize}
    ]
        \node[bnode] (A)  at (0.8,2.6) {$[\,]$};
        \node[bnode] (A1) at (0.8,1.3) {$(\,)$};
        \node[bnode] (B)  at (3.2,2.6) {$(\,)$};
        \node[bnode] (B1) at (2.5,1.3) {$\{\,\}$};
        \node[bnode] (B2) at (3.9,1.3) {$(\,)$};
        \draw[edge] (A) -- (A1);
        \draw[edge] (B) -- (B1);
        \draw[edge] (B) -- (B2);

        \node[chr] (c1)  at (0.30,0.0) {[};
        \node[chr] (c2)  at (0.75,0.0) {(};
        \node[chr] (c3)  at (1.20,0.0) {)};
        \node[chr] (c4)  at (1.65,0.0) {]};
        \node[chr] (c5)  at (2.30,0.0) {(};
        \node[chr] (c6)  at (2.75,0.0) {\{};
        \node[chr] (c7)  at (3.20,0.0) {\}};
        \node[chr] (c8)  at (3.65,0.0) {(};
        \node[chr] (c9)  at (4.10,0.0) {)};
        \node[chr] (c10) at (4.55,0.0) {)};

        \draw[link] (c1.north) to[bend left=22]  (A.south west);
        \draw[link] (c2.north) -- (A1.south);
        \draw[link] (c5.north) to[bend left=44] (B.west);
        \draw[link] (c6.north) -- (B1.south);
        \draw[link] (c8.north) -- (B2.south);

    \end{tikzpicture}
    }
    \captionsetup{font=footnotesize}
    \caption{A Dyck sentence is formed by well-balanced brackets and corresponds to a forest of trees, with inner brackets descending from outer ones. Dyck languages are a central object in theoretical linguistics for their ability to represent arbitrary depths of recursion.}
    \label{fig:dyck}
\end{wrapfigure}

\Glspl{llm} have demonstrated remarkable step-by-step reasoning capabilities for tasks ranging from mathematical word problems to coding, spurring intense research into understanding and improving procedural reasoning, \ie, the ability to solve problems systematically by following sound procedures. Inference-time techniques such as \gls{cot} consistently enhance accuracy on mathematical, logical, and code reasoning tasks \citep{Wei2022ChainOfThought}. However, current frontier models fail on sufficiently complex instances of computational tasks \citep{Shojaee2025IllusionThinking}, suggesting they never learn truly generalizable algorithms. Whether this reflects inherent architectural limitations or practical constraints of context length and compute budget is still contested \citep{Lawsen2025IllusionIllusion, Khan2025CommentIllusion, Varela2025RethinkingIllusion}, but mechanistic evidence also suggests that models achieve algorithmic behavior through ad-hoc ``bags of heuristics'', \ie, collections of narrow rules each firing on specific inputs \citep{Nikankin2024BagOfHeuristics}. Other works suggest that much of current \gls{cot} amounts to post hoc justification rather than a priori deliberation \citep{Turpin2023Unfaithful, Lanham2023Faithfulness}. Settling those and other disputes on frontier \glspl{llm} is hard due to opaque training data and the sheer computational cost of training them.

We sidestep those obstacles by moving to the opposite end of the scale. Tiny transformers have proven useful ``model organisms'' for isolating mechanisms that frontier models hopelessly entangle \citep{Nanda2023Progress}. Formal languages supply the graded, verifiable challenge such models need: because we can easily verify membership in simple formal languages and control their structure, we can generate inputs of any desired difficulty and verify results exactly \citep{Strobl2024FormalLanguageSurvey, Vafa2024WMImplicitInGM}. Concretely, we build two tasks over the natural bijection between forests of trees and Dyck-$k$ languages (sentences with up to $k$ bracket-types, correctly nested, see \Cref{fig:dyck}). The tasks admit protoreasoning traces: deterministic scratchpads that repeatedly prune tree nodes that cannot belong to the answer.

Our contributions are:
\begin{itemize}
  \item We induce a primitive but useful form of step-by-step reasoning in models far below the parameter scale required for natural-language competence, which can serve as minimal systems for studying reasoning.
  \item Based on the bijection between Dyck-$k$ languages and forests of trees, we propose two reasoning-friendly tasks with gradable difficulty and exact verifiability.
  \item We leverage the small size of our models to perform dense experimental designs, comprising over 1700 training runs.
\end{itemize}

\section{Methodology}
\label{sec:methodology}
\subsection{Tasks}
\label{sec:tasks}

In order to allow our models to display something akin to reasoning, we had to design tasks that were simple enough to be feasible given their small size, but challenging enough to benefit from step-by-step progression. Traditional, natural-language \gls{cot} was out of the question, for our models are much below the parameter count of even first-generation transformers \citep{Vaswani2017Attention}. Instead, we focused on simple formal languages and a latent syntax–semantics correspondence, then defined a couple of tasks, each with output given by a deterministic function of the input. The challenge we pose to the models is not sampling strings from the formal language per se, but providing the single correct answer to each input prompt belonging to the language.

\begin{wrapfigure}{r}{0.33\textwidth}
    \centering
    \resizebox{\linewidth}{!}{
    \begin{tikzpicture}[
        every node/.style={inner sep=0pt},
        bnode/.style={circle, draw, minimum size=8mm, font=\footnotesize},
        dp/.style={fill=taskdp!15},
        pl/.style={fill=taskpl!15},
        edge/.style={thick},
        linkdp/.style={-{Stealth[length=1.6mm]}, taskdp, densely dotted, semithick, shorten >=1pt, shorten <=1pt},
        linkpl/.style={-{Stealth[length=1.6mm]}, taskpl, densely dotted, semithick, shorten >=1pt, shorten <=1pt},
        chr/.style={font=\ttfamily\normalsize}
    ]
        \path[use as bounding box] (-0.55,-1.1) rectangle (5.2,4.65);
        \node[bnode, dp] (R)  at (2.1,4.2)  {$(\,)$};
        \node[bnode, dp] (N1) at (0.55,2.8) {$(\,)$};
        \node[bnode]     (N3) at (3.6,2.8)  {$\{\,\}$};
        \node[bnode, dp] (N2) at (0.55,1.4) {$[\,]$};
        \node[bnode, pl] (L3) at (2.6,1.4)  {$(\,)$};
        \node[bnode, pl] (L4) at (3.6,1.4)  {$[\,]$};
        \node[bnode, pl] (L5) at (4.6,1.4)  {$[\,]$};
        \node[bnode]     (L1) at (0.0,0.0)  {$\{\,\}$};
        \node[bnode, dp] (L2) at (1.1,0.0)  {$(\,)$};

        \draw[edge] (R) -- (N1);
        \draw[edge] (R) -- (N3);
        \draw[edge] (N1) -- (N2);
        \draw[edge] (N2) -- (L1);
        \draw[edge] (N2) -- (L2);
        \draw[edge] (N3) -- (L3);
        \draw[edge] (N3) -- (L4);
        \draw[edge] (N3) -- (L5);

        \node[chr, text=taskdp] (c1)  at (-0.40,-1) {(};
        \node[chr, text=taskdp] (c2)  at (-0.08,-1) {(};
        \node[chr, text=taskdp] (c3)  at ( 0.24,-1) {[};
        \node[chr, text=sp_token] (c4)  at ( 0.56,-1) {\{};
        \node[chr, text=sp_token] (c5)  at ( 0.88,-1) {\}};
        \node[chr, text=taskdp] (c6)  at ( 1.20,-1) {(};
        \node[chr, text=sp_token] (c7)  at ( 1.52,-1) {)};
        \node[chr, text=sp_token] (c8)  at ( 1.84,-1) {]};
        \node[chr, text=sp_token] (c9)  at ( 2.16,-1) {)};
        \node[chr, text=sp_token] (c10) at ( 2.48,-1) {\{};
        \node[chr, text=taskpl]  (c11) at ( 2.80,-1) {(};
        \node[chr, text=sp_token] (c12) at ( 3.12,-1) {)};
        \node[chr, text=taskpl]  (c13) at ( 3.44,-1) {[};
        \node[chr, text=sp_token] (c14) at ( 3.76,-1) {]};
        \node[chr, text=taskpl]  (c15) at ( 4.08,-1) {[};
        \node[chr, text=sp_token] (c16) at ( 4.40,-1) {]};
        \node[chr, text=sp_token] (c17) at ( 4.72,-1) {\}};
        \node[chr, text=sp_token] (c18) at ( 5.04,-1) {)};

        \draw[linkdp] (c1.north) to[bend left=45] (R.west);
        \draw[linkdp] (c2.north) to[bend left=35] (N1.south west);
        \draw[linkdp] (c3.north) to[bend right=15] (N2.south);
        \draw[linkdp] (c6.north) -- (L2.south);
        \draw[linkpl] (c11.north) -- (L3.south);
        \draw[linkpl] (c13.north) -- (L4.south);
        \draw[linkpl] (c15.north) -- (L5.south);
    \end{tikzpicture}
    }
    \captionsetup{font=footnotesize}
    \caption{Tasks. \textbf{Deepest path} (\gls{dp}, answer in {\color{taskdp}blue}): path to the deepest leaf. \textbf{Max}imum-order \textbf{preleaf} (\gls{pl}, answer in {\color{taskpl}orange}): largest all-leaf sibling brood. Ties are broken rightmost and the answer has only the opening brackets. Input Dyck sentence at the figure's bottom.}
    \label{fig:tasks}
\end{wrapfigure}
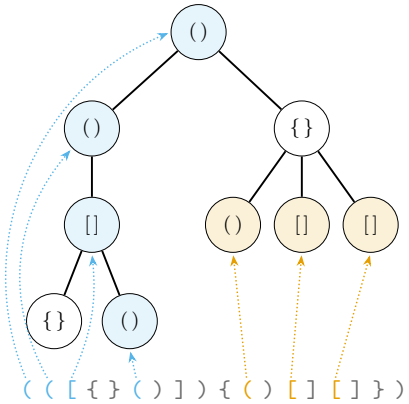

Dyck-$k$ languages are the sets of all sentences with $k$ types of brackets correctly nested and balanced. For example, $([\,])$ is a Dyck-$k$ sentence for $k\ge2$, but $[(\,])$ is not a Dyck sentence for any $k$. Theoretical linguistics has studied these languages extensively for their ability to represent arbitrary depths of grammatical recursion while remaining extremely simple.

Semantically, each Dyck sentence can be interpreted as a tree (or forest), with inner brackets representing descendants of embracing outer brackets. Using that correspondence, we define two tasks: deepest path (\gls{dp}), which outputs the opening brackets along the path that leads to the deepest leaf in the sentence, and maximum-order preleaf (\gls{pl}), which outputs the opening brackets of the biggest group of siblings composed entirely of leaves. In both tasks the expected answer holds only the \emph{opening} brackets. Ties are broken in favor of the rightmost candidate (\Cref{fig:tasks}).

We characterize each sentence by its \textbf{half-length} $\mathrm{hl}$ (number of bracket pairs in the sentence) and a \textbf{structural parameter} related to the task (depth or maximum preleaf order). Then, we explore generalization across those two axes.

\paragraph{Why tiny models?} Scale is crucial for \gls{llm} performance: arguably, the transformer decoder became the dominant architecture for language models because it scales up so well. Still, tiny \glspl{llm} have proven useful ``model organisms'' for exploring phenomena that appear in much bigger networks. State-of-the-art \glspl{llm} take weeks and require dozens of multi-GPU nodes to train. ``Small'' models (1 to 10B parameters) still can take days on an expensive node to pretrain. In contrast, our entire study, with thousands of training runs, takes little more than a day on a multi-GPU node. See \Cref{sec:app-details} for details on compute.

\subsection{Protoreasoning traces}
\label{sec:traces}
\begin{figure}[t]
    \centering
    \begin{minipage}[t]{0.48\textwidth}
        \centering
        \resizebox{\linewidth}{!}{
        \begin{tikzpicture}[baseline=(current bounding box.north)]
            \node[
            align=left,
            text width=6cm,
            font=\ttfamily
            ] {

            {\color{sp_token}<BOS>}~(([\{\}()])\{()[][]\})~{\color{taskdp}|}\\
            {\color{sp_token}<BOT>}\\
            {\color{trace}(([\{\}()])\{[][]\}) {\color{sp_token}$\hookleftarrow$}\\
                (([\{\}()])\{[]\}) {\color{sp_token}$\hookleftarrow$}\\
                (([\{\}()])\{\}) {\color{sp_token}$\hookleftarrow$}\\
                (([\{\}()])) {\color{sp_token}$\hookleftarrow$}\\
                (([()])) {\color{sp_token}$\hookleftarrow$}\\
            }
            {\color{sp_token}<EOT>}\\{\color{taskdp}(([(}~{\color{sp_token}<EOS>}

            };
        \end{tikzpicture}
        }
    \end{minipage}\hfill
    \begin{minipage}[t]{0.48\textwidth}
        \centering
        \resizebox{\linewidth}{!}{
        \begin{tikzpicture}[baseline=(current bounding box.north)]
            \node[
            align=left,
            text width=6cm,
            font=\ttfamily
            ] {

            {\color{sp_token}<BOS>}~(([\{\}()])\{()[][]\})~{\color{taskpl}!}\\
            {\color{sp_token}<BOT>}\\
            {\color{trace}([\{\}()])\{()[][]\} {\color{sp_token}$\hookleftarrow$}\\
                {}[\{\}()]\{()[][]\} {\color{sp_token}$\hookleftarrow$}\\
                \{()[][]\} {\color{sp_token}$\hookleftarrow$}\\
            }
            {\color{sp_token}<EOT>}\\{\color{taskpl}([[}~{\color{sp_token}<EOS>}

            };
        \end{tikzpicture}
        }
    \end{minipage}
    \caption{Training samples for input and tasks of \Cref{fig:tasks}. A token separates input and expected output: {\color{taskdp}|} for \gls{dp} (left) and {\color{taskpl}!} for \gls{pl} (right). Each step of the {\color{trace}reasoning trace} prunes part of the trees until only the answer remains. The {\color{sp_token}$\hookleftarrow$} marks newline tokens between steps: all other spacing is just for visualization. Special {\color{sp_token}<tokens>} delimit the sentence and the trace. Vanilla samples omit the trace and its delimiters.}
    \label{fig:trace}
\end{figure}

The main experimental intervention is adding an explicit reasoning trace to training samples, which solves the task step-by-step by eliminating brackets/tree nodes that cannot be part of the solution. Thus, we propose every task in two formats: a \textbf{vanilla} training sample has, in sequence, the input sentence, the task separator, and the expected output; a \textbf{trace} training sample adds a reasoning trace $s_1,\dots,s_m$ (\Cref{fig:trace}). Each step $s_i$ is the previous step $s_{i-1}$ after pruning some part of the tree that cannot be part of the solution.

For \gls{dp}, at each step, we eliminate the shallowest leaf still present, until the tree degenerates into a path graph. For \gls{pl}, we peel the trees by the root, and whenever two or more trees have only two levels, we eliminate the smallest one, until only a single tree remains.\footnote{As a special case, if the input is a forest of single nodes (\eg, $\{\}\{\}\{\}$), we consider that forest the largest sibling brood.} In both cases, if there are tied candidates for elimination, we prune the leftmost.

To make the traces more challenging and to stimulate generalization, we introduce a \textbf{step dropout}, which randomly skips reasoning steps.

The traces are never part of the prompts we use for evaluation. At test time, both formats are prompted with the sample prefix up to and including the task separator.

\subsection{Experimental setting}
\label{sec:settings}

Unless explicitly stated otherwise, we use:
\begin{itemize}
    \item Dyck-4 languages with sentences of 32 bracket pairs (half-length $\mathrm{hl}=32$).
    \item A miniaturized Llama 2 model with 4 layers, residual width 128, 16 heads, and no grouped-query attention.
    \item 10 random seeds affecting model-weight initialization and training-data sampling.
    \item An 8-trial quasi-random hyperparameter search per treatment. For each treatment, hyperparameter search runs only once (for the first random seed).
    \item A positional-encoding data augmentation (SkipAlign, see below).
    \item For trace treatments, step dropout with a drop probability of 15\%, which we apply independently for each reasoning step.
\end{itemize}

Because we are operating close to the limit of the capacity of the models (see \Cref{sec:app-details}), the experiments can be noisy, with models sometimes alternating between total failure and perfect success depending on how we initialize weights and data sampling. To cope with such noise, we average all reported numbers over the 10 random seeds.

\paragraph{Hold-out patterns.}
Each value of the structural parameter is a data \textbf{stratum} in the design. We probe generalization by splitting the strata into disjoint training, validation, and test sets, using six different patterns, named by which strata are held out (validation+test): \textbf{R}andom, two \textbf{B}locks, \textbf{E}ven, \textbf{M}iddle, \textbf{H}igh, and \textbf{L}ow. R/B/E/M emphasize interpolation (short-range for R/E, mid-range for B, long-range for M), while H/L emphasize extrapolation. The exact splits we employed appear in \Cref{fig:pattern-atlas} in the Appendix, and are also visible in the styling of the x-axes' ticks in \Cref{fig:gen-pure} and \Cref{fig:gen-mixed}.

\paragraph{Metrics.}
The main metric is output \textbf{validity}: whether the output is correct given the input. We take as output anything after the \texttt{<EOT>} token (or the task separator, if \texttt{<EOT>} is not present) and before the \texttt{<EOS>} token. At every step, we draw fresh sentences from a generator so large that, for all practical purposes, we may consider that the model never trains on the same sentence twice.
Thus, the validity on each new training batch already measures in-domain generalization (new sentences on already seen train strata). However, unless we explicitly state otherwise, we measure and report \textbf{out-of-distribution/OOD generalization} averaged over the unseen test strata.

\paragraph{Deconfounding positional-encoding generalization.}
The length of any sample depends on both the half-length and structural parameter of the input, but the range of lengths is much wider for trace samples. Using standard \gls{rope}, trace pays a big penalty because we test it on positional encodings that differ substantially from those it saw in training. To eliminate this confounder, we train all models with \textbf{SkipAlign}~\citep{Wu2024SkipAlign}, a PoSE-style~\citep{Zhu2024PoSE} positional augmentation that inserts position-index gaps at strategic boundaries. SkipAlign is a form of data augmentation that allows training the model with much more diverse positional encodings, without paying the quadratic cost of using large contexts filled with actual data.

More details, including hyperparameter search, training, and data sampling, appear in the Appendix.

\section{Experiments Design and Results}
\label{sec:results}
We summarize our main findings below. The corresponding subsections report details and further analyses.

\paragraph{The proposed tasks are easy to express but hard to generalize.} Our tasks strike the right level of challenge for the models we evaluate (\Cref{sec:res-expr}). A model trained on a single stratum reaches near-perfect validity when tested in-distribution, but collapses out-of-distribution, especially along the structural axis (\Cref{tab:expr}).

\paragraph{Protoreasoning traces improve generalization.} When we instead train on diverse strata, out-of-distribution generalization markedly improves. Across two tasks and six hold-out patterns, the trace format almost always beats the vanilla one (\Cref{sec:res-patterns}, \Cref{tab:gen-pure}).

\paragraph{Protoreasoning traces help mixed tasks.} The advantage of protoreasoning traces increases when we train a single model on both tasks at once: trace tolerates the multi-task setting far better than vanilla (\Cref{sec:res-mixed}, \Cref{tab:gen-mixed}).

\paragraph{Trace's improvement stems from its contents, not simply the extra tokens it adds.} Replacing trace content with filler dots collapses the gains (\Cref{tab:trace-ablation}, \Cref{sec:res-ablations}).

\begin{table}[!htbp]
  \centering
  \small
  \caption{Expressivity versus generalization for models trained on a single stratum of the vanilla format. Output validity (\%) averaged over the concerned strata and 10 random seeds $\pm$ standard deviation over the seeds, for three training half-lengths (hl). In-domain validity is near perfect, but generalization crashes out-of-distribution, especially along the structural axis (depth/max. preleaf).}
  \label{tab:expr}
  \begin{btvtable}{C C V C C C C}
\textsc{Train HL} & \textsc{Train Struct.} & \textsc{In-Domain} & \textsc{HL OOD} & \textsc{Struct. OOD} & \textsc{Both OOD} \\
\btvheadrule
\SetCell[c=6]{c} \textsc{Deepest Path} & & & & & \\
\hline
32  & 16 & $99.8 \pm 0.5$ & $34.9 \pm 13.5$ & $0.0 \pm 0.0$ & $0.2 \pm 0.2$ \\
64  & 32 & $99.8 \pm 0.4$ & $4.4 \pm 4.7$   & $0.0 \pm 0.0$ & $0.3 \pm 0.4$ \\
128 & 64 & $98.4 \pm 1.8$ & $1.8 \pm 1.4$   & $0.0 \pm 0.0$ & $0.0 \pm 0.0$ \\
\hline
\SetCell[c=6]{c} \textsc{Max. Preleaf} & & & & & \\
\hline
32  & 8  & $99.7 \pm 0.4$ & $62.6 \pm 28.2$ & $0.0 \pm 0.0$ & $0.5 \pm 0.9$ \\
64  & 16 & $99.8 \pm 0.3$ & $91.1 \pm 10.3$ & $0.2 \pm 0.5$ & $0.6 \pm 1.5$ \\
128 & 32 & $99.7 \pm 0.4$ & $50.1 \pm 7.4$  & $0.0 \pm 0.0$ & $0.0 \pm 0.0$ \\
\end{btvtable}

\end{table}

\begin{table}[!htbp]
  \centering
  \small
  \caption{Generalization across different holdout patterns for models trained and evaluated on a single task (\glsentrylong{dp}/\glsentryshort{dp} or \glsentrylong{pl}/\glsentryshort{pl}) and format (vanilla or trace). Output validity averaged over the test strata and 10 random seeds $\pm$ standard deviation over the seeds. Reasoning traces almost always improve generalization.}
  \label{tab:gen-pure}
  \begin{btvtable}{C V C C V C C}
 & \SetCell[c=2]{c} \textsc{Deepest Path} &  & \SetCell[c=2]{c} \textsc{Max. Preleaf} &  \\
\textsc{Pattern} & \textsc{Vanilla} & \textsc{Trace} & \textsc{Vanilla} & \textsc{Trace} \\
\btvheadrule
Random & $93.6 \pm 4.5$          & $\mathbf{95.4 \pm 2.1}$ & $91.7 \pm 2.4$          & $\mathbf{97.8 \pm 0.5}$ \\
Blocks & $94.6 \pm 2.4$          & $\mathbf{98.7 \pm 0.4}$ & $76.9 \pm 7.0$          & $\mathbf{97.3 \pm 0.5}$ \\
Even   & $98.5 \pm 2.5$          & $\mathbf{98.8 \pm 0.9}$ & $88.4 \pm 0.8$          & $\mathbf{97.7 \pm 0.5}$ \\
Middle & $80.5 \pm 9.7$          & $\mathbf{98.0 \pm 2.0}$ & $85.8 \pm 11.0$         & $\mathbf{97.1 \pm 1.1}$ \\
High   & $50.9 \pm 5.4$          & $\mathbf{66.7 \pm 9.4}$ & $23.9 \pm 24.3$         & $\mathbf{62.2 \pm 7.3}$ \\
Low    & $\mathbf{39.6 \pm 2.0}$ & $21.6 \pm 7.4$          & $\mathbf{43.9 \pm 3.0}$ & $41.0 \pm 5.4$          \\
\end{btvtable}

\end{table}

\begin{table}[!htbp]
  \centering
  \small
  \caption{Generalization across different holdout patterns for models trained on both tasks (\gls{dp} and \gls{pl}) and evaluated on each of them. Output validity averaged over the test strata and 10 random seeds $\pm$ standard deviation over the seeds. In general, trace tolerates multi-task training much better than vanilla.}
  \label{tab:gen-mixed}
  \begin{btvtable}{C V C C V C C}
 & \SetCell[c=2]{c} \textsc{Evaluated on DP} &  & \SetCell[c=2]{c} \textsc{Evaluated on PL} &  \\
\textsc{Pattern} & \textsc{vanilla} & \textsc{trace} & \textsc{vanilla} & \textsc{trace} \\
\btvheadrule
Even Mixed   & $62.7 \pm 33.2$         & $\mathbf{98.6 \pm 0.2}$ & $74.4 \pm 15.8$ & $\mathbf{94.5 \pm 1.6}$ \\
Middle Mixed & $80.7 \pm 16.8$         & $\mathbf{92.7 \pm 2.9}$ & $60.5 \pm 24.1$ & $\mathbf{94.8 \pm 1.7}$ \\
High Mixed   & $\mathbf{46.6 \pm 5.2}$ & $43.0 \pm 4.9$          & $17.7 \pm 13.7$ & $\mathbf{60.5 \pm 7.7}$ \\
\end{btvtable}

\end{table}

\begin{table}[!htbp]
  \centering
  \small
  \caption{Ablations across holdout patterns. Output validity averaged over the test strata and 10 random seeds $\pm$ standard deviation over the seeds. Both data augmentations (SkipAlign and step dropout) help improve generalization, especially on deepest path, which has longer reasoning traces. Erasing the trace content with dots dramatically deteriorates results, showing that increasing the context size does not suffice to explain trace's contribution.}
  \label{tab:trace-ablation}
  \begin{btvtable}{C D C C C C V C D C C}
 & \SetCell[c=5]{c} \textsc{Trace} &  &  &  &  & \SetCell[c=2]{c} \textsc{Vanilla} &  \\
\cline[leftpos=0, rightpos=0, endpos]{2-6}\cline[leftpos=0, rightpos=0, endpos]{7-8}
\textsc{Pattern} & \SetCell[c=2]{c} \textsc{All steps} &  & \SetCell[c=2]{c} $+$ \textsc{Step dropout} &  & \SetCell[r=2]{c} \textsc{Erased} & \SetCell[r=2]{c} \textsc{RoPE} & \SetCell[r=2]{c} \textsc{SkipAlign} \\
 & \textsc{RoPE} & \textsc{SkipAlign} & \textsc{RoPE} & \textsc{SkipAlign} &  &  &  \\
\btvheadrule
\SetCell[c=8]{c} \textsc{Deepest Path} & & & & & & & \\
\hline
Random & $59.8 \pm 24.4$         & $85.2 \pm 8.5$          & $94.4 \pm 1.7$          & $\mathbf{95.4 \pm 2.1}$ & $9.8 \pm 9.0$   & $88.9 \pm 3.8$           & $\mathbf{93.6 \pm 4.5}$  \\
Blocks & $81.1 \pm 15.2$         & $94.3 \pm 7.6$          & $\mathbf{99.2 \pm 0.5}$ & $98.7 \pm 0.4$          & $5.4 \pm 9.0$   & $\mathbf{94.9 \pm 2.5}$  & $94.6 \pm 2.4$           \\
Even   & $64.6 \pm 38.7$         & $91.7 \pm 20.1$         & $\mathbf{99.1 \pm 0.3}$ & $98.8 \pm 0.9$          & $0.5 \pm 0.8$   & $95.7 \pm 10.6$          & $\mathbf{98.5 \pm 2.5}$  \\
Middle & $87.8 \pm 23.0$         & $93.4 \pm 11.4$         & $87.5 \pm 10.2$         & $\mathbf{98.0 \pm 2.0}$ & $4.3 \pm 6.7$   & $76.3 \pm 5.9$           & $\mathbf{80.5 \pm 9.7}$  \\
High   & $43.5 \pm 16.0$         & $54.7 \pm 9.6$          & $53.5 \pm 3.0$          & $\mathbf{66.7 \pm 9.4}$ & $0.0 \pm 0.0$   & $49.3 \pm 3.7$           & $\mathbf{50.9 \pm 5.4}$  \\
Low    & $3.5 \pm 3.7$           & $18.7 \pm 11.9$         & $10.9 \pm 8.1$          & $\mathbf{21.6 \pm 7.4}$ & $0.0 \pm 0.0$   & $34.4 \pm 3.6$           & $\mathbf{39.6 \pm 2.0}$  \\
\hline
\SetCell[c=8]{c} \textsc{Max. Preleaf} & & & & & & & \\
\hline
Random & $98.1 \pm 1.9$          & $\mathbf{98.7 \pm 0.9}$ & $97.8 \pm 0.9$          & $97.8 \pm 0.5$          & $30.3 \pm 12.9$ & $84.1 \pm 6.2$           & $\mathbf{91.7 \pm 2.4}$  \\
Blocks & $\mathbf{97.5 \pm 0.6}$ & $97.2 \pm 0.9$          & $97.1 \pm 0.6$          & $97.3 \pm 0.5$          & $44.5 \pm 5.3$  & $\mathbf{79.8 \pm 10.5}$ & $76.9 \pm 7.0$           \\
Even   & $95.8 \pm 3.6$          & $97.5 \pm 3.0$          & $96.4 \pm 2.0$          & $\mathbf{97.7 \pm 0.5}$ & $36.4 \pm 13.7$ & $82.3 \pm 12.8$          & $\mathbf{88.4 \pm 0.8}$  \\
Middle & $94.5 \pm 3.8$          & $\mathbf{97.5 \pm 1.4}$ & $96.7 \pm 1.8$          & $97.1 \pm 1.1$          & $29.1 \pm 14.3$ & $79.9 \pm 15.4$          & $\mathbf{85.8 \pm 11.0}$ \\
High   & $48.9 \pm 4.4$          & $58.1 \pm 9.6$          & $56.6 \pm 8.6$          & $\mathbf{62.2 \pm 7.3}$ & $0.0 \pm 0.0$   & $19.7 \pm 16.7$          & $\mathbf{23.9 \pm 24.3}$ \\
Low    & $35.7 \pm 9.3$          & $\mathbf{47.6 \pm 3.9}$ & $34.6 \pm 9.6$          & $41.0 \pm 5.4$          & $11.3 \pm 7.3$  & $\mathbf{44.0 \pm 3.4}$  & $43.9 \pm 3.0$           \\
\end{btvtable}

\end{table}

\begin{figure}[!htbp]
  \centering
  \includegraphics[width=\textwidth]{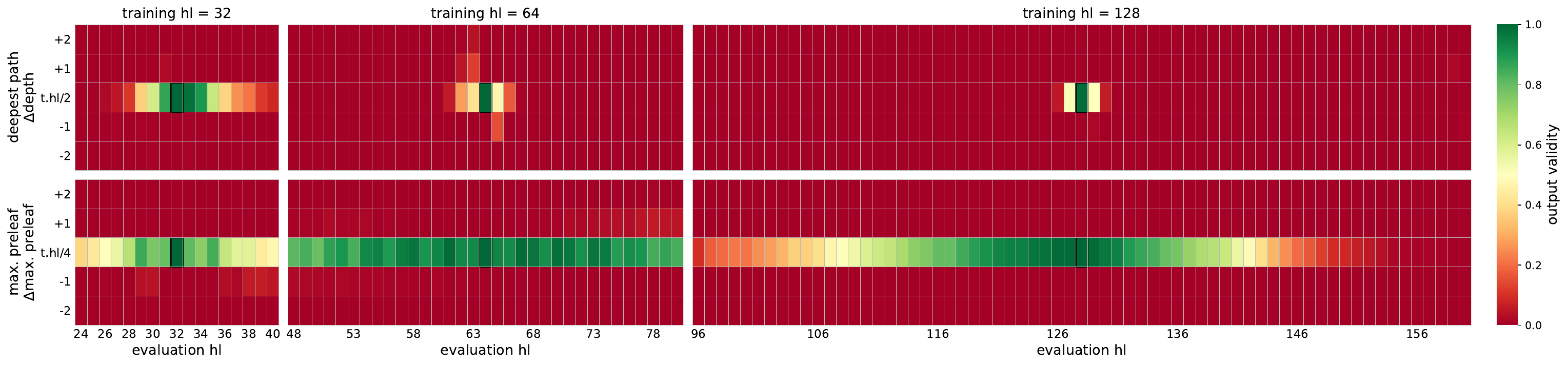}
  \caption{Six experiments showcasing the gap between expressivity and generalization for the tasks proposed. We train the models with a single stratum, with in-distribution validity appearing as the dark-green square at the center of each subplot, showing the models are capable of expressing the task. All other squares in the subplot are out-of-distribution (hl and structure offsets shown horizontally and vertically, respectively). Both length-wise and structure-wise generalizations are hard, but the latter is especially difficult.}
  \label{fig:anchor-grids}
\end{figure}

\subsection{The proposed tasks are easy to express but hard to generalize}
\label{sec:res-expr}

All experiments in this section use the vanilla sample format, and vary half-lengths and structural parameters.

\Cref{tab:expr} showcases the tension between expressivity and generalization.
Near-perfect validity shows that the model has enough capacity to express the correct solution for in-distribution prompts, even for over a hundred brackets and very complex structure. However, the picture reverses out-of-distribution: validity is only partly retained along the length axis (hl), and all but vanishes along the structural axis (depth for \gls{dp}, maximum preleaf for \gls{pl}).

\Cref{fig:anchor-grids} shows the same results broken down by test stratum. The dark-green square in the middle of each subplot shows the in-distribution validity (always $\sim$100\%). All other squares are out-of-distribution. Pure-length generalization strata are in the same row as the in-distribution square, while pure-structural generalization strata share the same column. It is immediately visible that the former are easier than the latter, although both remain challenging.

\begin{figure}[!htbp]
  \centering
  \scalebox{1}[0.91]{\includegraphics[width=\textwidth]{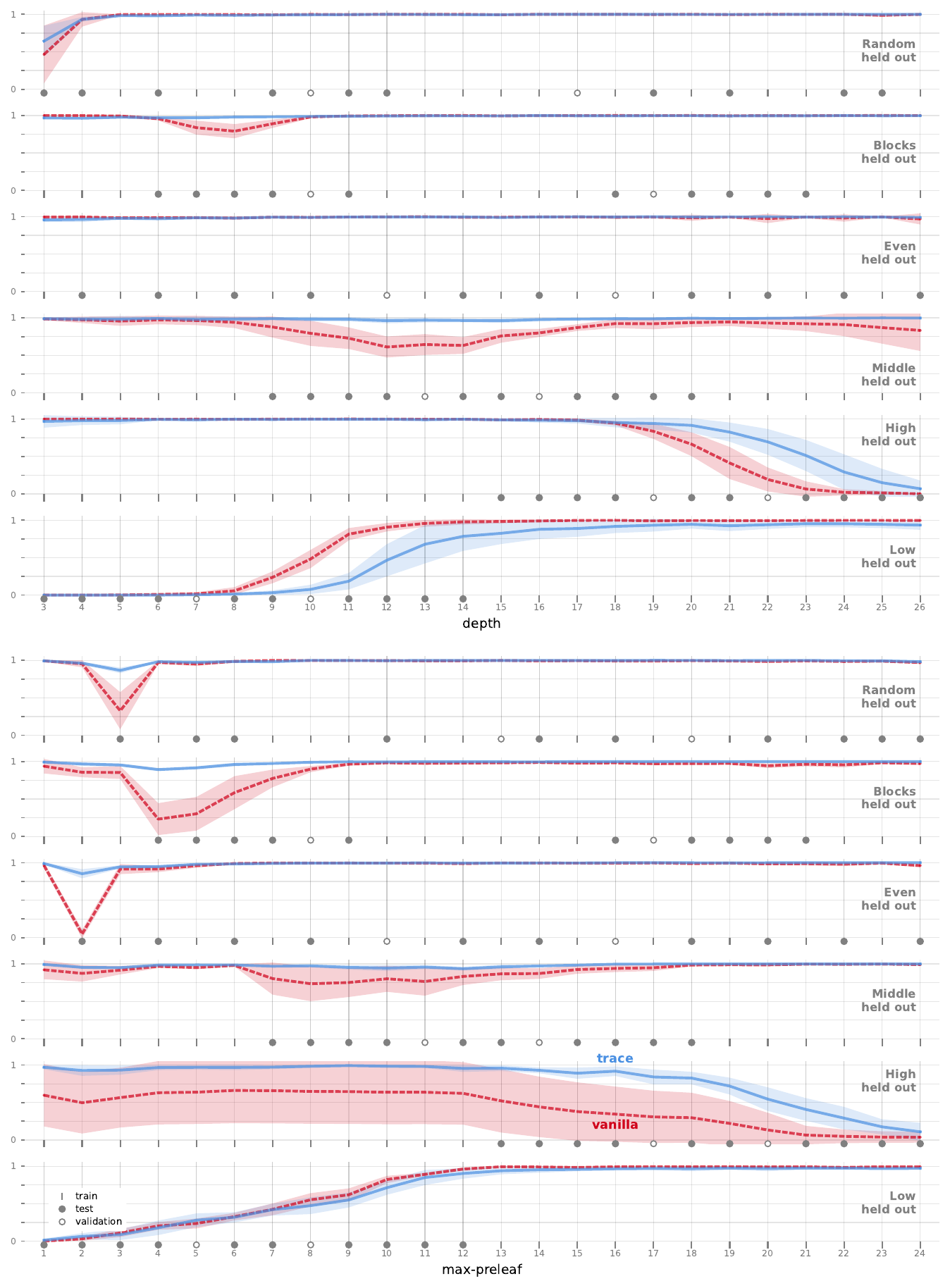}}
  \caption{Same results as in \Cref{tab:gen-pure}, shown per stratum. Averaged output validities (lines) $\pm$ standard deviations (shaded areas) over 10 random seeds. Top: \glsforcefull{dp}, bottom: \glsforcefull{pl}. Comparison between vanilla and trace formats for each pattern $\times$ task combination. Training, validation, and test strata are shown in the x-axes' tick shapes. In general, reasoning traces handle long interpolations and extrapolations better, although extrapolation remains hard for both formats.}
  \label{fig:gen-pure}
\end{figure}

\subsection{Protoreasoning traces help generalization, but extrapolation is still a challenge}
\label{sec:res-patterns}

The previous section showed that we should not expect zero-shot generalization when training with a single stratum, especially along the structural axis. In this section, we show that generalization is still possible if we train the model with a diverse array of strata. Thus, we split the strata into train, validation, and test using the holdout patterns explained in \Cref{sec:settings}.

\Cref{tab:gen-pure} shows that the trace format generalizes better than vanilla almost always across holdout patterns, often by a considerable margin. On the interpolation patterns (R, B, E, M) the trace format leads to validities above 95\%, with small deviation across seeds. The extrapolation regime (H, L), however, is hard for both formats: validity drops precipitously and becomes much noisier. Pattern L is particularly challenging for trace, because it leads to holdout reasoning traces having many more steps than the ones seen in training.

\Cref{fig:gen-pure}, with results broken down by stratum, makes the generalization differences more evident. The in-domain strata used for training (straight x-axis ticks) have near-perfect validity, while the held-out out-of-domain squares (round ticks) may have anything from perfect to zero validity, depending on the hold-out pattern, data format, and specific stratum. In the interpolation regime (R, B, E, M), trace always beats vanilla, especially on the longer-range interpolation of B and M. Neither trace nor vanilla is able to extrapolate very far.

\begin{figure}[!htbp]
  \centering
  \includegraphics[width=\textwidth]{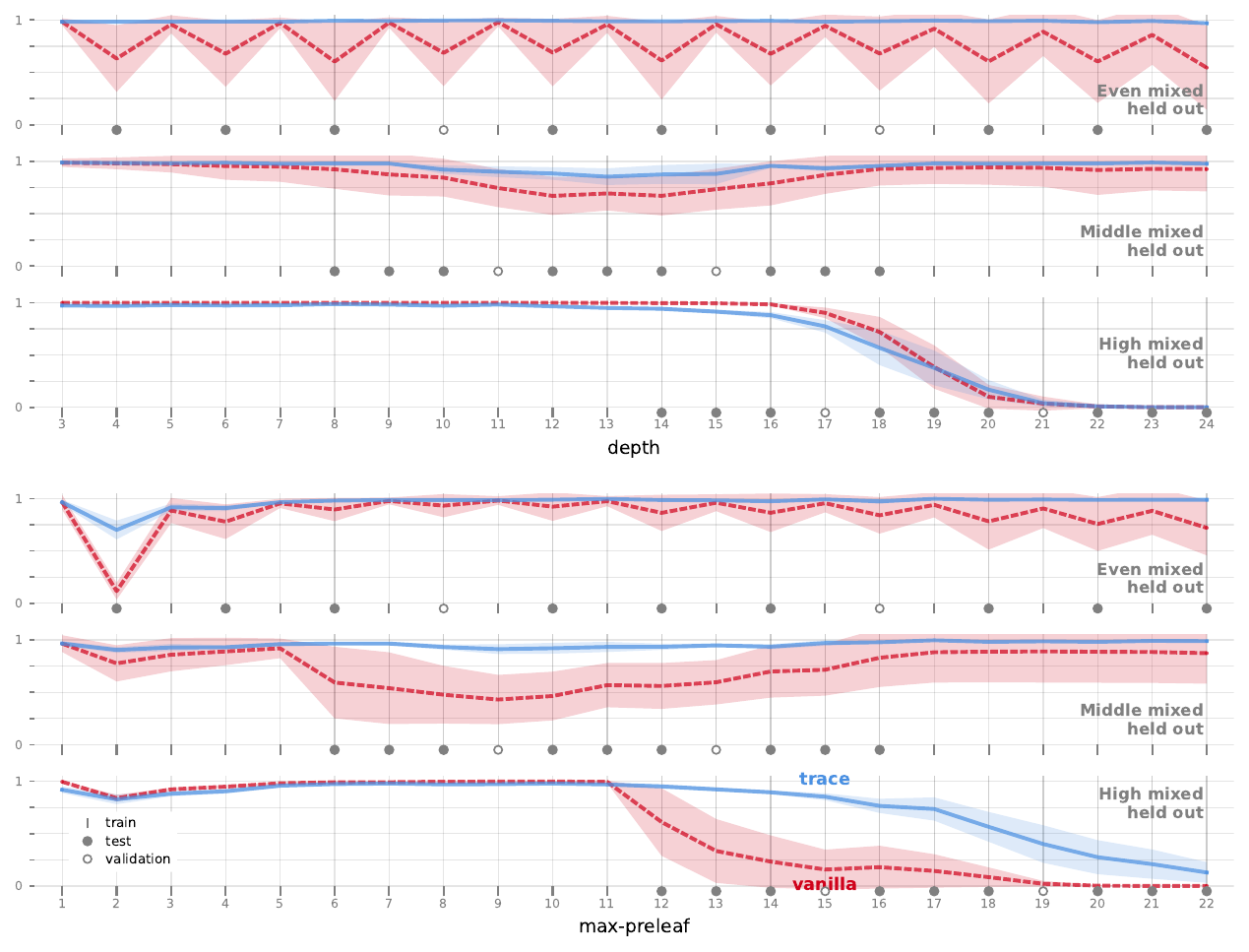}
  \caption{Same results as in \Cref{tab:gen-mixed}, shown per stratum. Averaged output validities (lines) $\pm$ standard deviations (shaded areas) over 10 random seeds. All models trained on both tasks and evaluated separately on \glsentrylong{dp}/\glsentryshort{dp} (top) and \glsentrylong{pl}/\glsentryshort{pl} (bottom). Training, validation, and test strata are shown in the x-axes' tick shapes. Trace preserves generalization better under mixed-task training than vanilla.}
  \label{fig:gen-mixed}
\end{figure}

\subsection{Protoreasoning traces help mixed tasks}
\label{sec:res-mixed}

Here, we train all models on both tasks (\gls{dp} and \gls{pl}) and then evaluate each task separately. We expect the models to provide the right output for the right task, using the input separators | or ! to identify whether the desired output is for \gls{dp} or \gls{pl}, respectively.

We stratify by the two structural parameters, depth and maximum pre-leaf, at once. Because the parameters interact (for a fixed number of nodes, a tree cannot, at once, be very deep and have a very large preleaf) we sample only from the combinations that have enough diversity so that the model does not just trivially memorize them. As usual, we sample a stratum uniformly at random, and then a sentence within the stratum uniformly at random, but now we have many more strata, reflecting the joint choice of parameters (\Cref{sec:app-patterns}).

To evaluate out-of-distribution generalization, we adapt patterns E, M, and H to consider only the joint strata explained above. We apply the hold-out patterns independently for each task: \gls{dp} samples apply the holdout pattern to depth, and \gls{pl} samples to maximum preleaf. The independent stratification ensures that the model sees all possible input strata, but for each task separately, the train/test/validation split still applies.

\Cref{tab:gen-mixed} shows that mixed-task training is more challenging for both formats, but that validity drops much more for vanilla than for trace. \Cref{fig:gen-mixed} details this per stratum, where the gap is most visible on the longer-range interpolative patterns H and M. We hoped that some form of cross-transfer could happen, \ie, that one task would be able to better generalize to its held-out strata by noticing that those still appear for the other task, but that would have led to \textit{improved} results in contrast to the pure tasks of \Cref{sec:res-patterns}, which was not the case. That suggests that generalization challenges are due to how the model wires the solution for each task, and not simply to the model being unable to represent a given input.

\subsection{Ablations}
\label{sec:res-ablations}

To help the model generalize to new strata, we use two data augmentations. Both formats (vanilla and trace) employ SkipAlign, which introduces gaps on the index used to compute the positional encoding. Trace, additionally, employs step dropout, which randomly skips 15\% of the reasoning steps.

\Cref{tab:trace-ablation} shows that SkipAlign improves results on both data formats and both tasks. Step dropout has a large positive effect on deepest path, which has long reasoning traces and a lot of length variability over strata. Although it slightly penalizes maximum preleaf, whose traces are much shorter, its average effect is still positive.

A possible confounder for the improvements brought by trace is the sheer addition of more tokens. Indeed, literature shows that additional non-informative tokens may improve performance simply by allowing the model a larger compute budget \citep{Pfau2024DotByDot}. To exclude that possibility, we implement a ``dot-by-dot'' replacement of the reasoning traces at training time, which dramatically deteriorates validity, showing that the trace contents indeed contribute to the validity improvements.

\section{Related Work}
\label{sec:related}
\paragraph{Step-by-step reasoning.}
\citet{Wei2022ChainOfThought} showed that simply prompting a large model to emit its intermediate steps, a technique that became known as \gls{cot}, markedly improves accuracy on arithmetic, commonsense, and symbolic reasoning, triggering a wave of work on step-by-step reasoning. A substantial survey literature now maps that landscape \citep{Huang2023SurveyTowardsReasoning, Plaat2024SurveyMultistep, Chu2024SurveyCoT, Sun2025SurveyReasoningFM, Li2025SurveySystem2}. Its recurring finding is that eliciting intermediate steps (\gls{cot} prompting, scratchpads, self-consistency, and tree or graph search) reliably improves accuracy on mathematical, logical, and code tasks, though the gains saturate as problem complexity grows. Tool integration \citep{Schick2023Toolformer, Yao2023ReAct} and structured search \citep{Yao2023TreeOfThoughts, Besta2024GraphOfThoughts} further enhance and stabilize performance.

\paragraph{The limits of reasoning.}
Even with \gls{cot}, models remain brittle out-of-distribution \citep{Mondorf2024SurveyBeyondAcc}. \citet{Shojaee2025IllusionThinking} report that, for sufficiently complex instances of computational puzzles (Tower of Hanoi, river crossing, etc.), both reasoning and non-reasoning models eventually fail, with accuracy collapsing past a complexity threshold. Others contest whether this reflects an inherent limitation or an artifact of the evaluation: bounded token budgets, disallowed tools, unsatisfiable instances, or crediting only the strictly optimal solution \citep{Lawsen2025IllusionIllusion, Khan2025CommentIllusion, Varela2025RethinkingIllusion}. Another limitation is that reasoning traces are not necessarily faithful: the emitted steps need not reflect the computation that actually produced the answer \citep{Turpin2023Unfaithful, Lanham2023Faithfulness}.

\paragraph{Formal languages.}
Formal languages apply the rigorous lens of theoretical computer science to this investigation. The well-established Chomsky hierarchy of formal languages (regular, context-free, context-sensitive, and recursively enumerable), as well as a finer-grained web of languages characterized by the complexity class of the decision algorithm required to recognize them, allows us to gauge a model's capability by asking which of those language classes it can generate or recognize.

\citet{Strobl2024FormalLanguageSurvey} survey that literature, reporting bounds that place transformer-based models above (``at least''), below (``at most''), or in equivalence with complexity classes, and noting that the conclusions hinge on modeling assumptions (architecture, positional encoding, numerical precision, attention masking/activation) that often assume strong idealizations (infinite precision, hard attention). Under relatively realistic assumptions, some analyses \citep{Chiang2023TighterBounds, Merrill2023LogPrecision, Lin2021LimitationsAutoregressive, Hahn2020TheoreticalLimitations} yield informative results: encoder-only models can express the languages in uniform-ACC$^0$ under constant precision, extending to uniform-TC$^0$ under $O(\log n)$ precision; decoder-only models are less well understood, but known counterexamples show they cannot recognize Parity or Dyck-2. These guarantees, however, speak to expressivity rather than learnability, \ie, whether a model can represent a language, not whether it will learn it from data.

Such limitations, which also include a focus on asymptotic analysis, motivate a more empirical approach, which employs real-world architectures without extra assumptions and measures learning from data. \citet{Vafa2024WMImplicitInGM} measure, empirically, how GPT-2-like models learn a regular automaton (conceptualized as a world model), showing that next-token accuracy is a poor proxy for learning the true automaton: the \gls{llm} may display excellent next-token prediction and still break down completely under small deviations, demonstrating that a robust model of the automaton was not learned.

 \citet{AllenZhu2023Physics} pre-train GPT-2-style decoders on synthetic \glspl{cfg} conceived to be long and locally ambiguous, so that deciding membership requires global disambiguation, finding that the models generate valid outputs and encode in their hidden states the grammar's nonterminals, while their attention patterns echo the information flow of a dynamic-programming parser. \citet{Nandakumar2026Expressivity} supply matching results for bounded-depth, non-recursive \glspl{cfg}, explicitly constructing transformers whose depth grows linearly with grammar depth and that map each nonterminal to a low-dimensional, linearly separable subspace of the residual stream. Learnability nonetheless stays fragile with depth: \citet{Li2026DiagnosingCFG}, probing models as in-context interpreters of novel \glspl{cfg}, report that structural semantics may collapse under deep recursion or heavy branching, even while surface syntax still survives. Using invented words, they expose a reliance on the semantic priors of familiar keywords rather than genuine symbolic induction.
 \citet{Ghosh2026FormalBenchmark} advocates probabilistic formal languages as semantic-contamination-free testbeds, a framework used later to pit fine-tuning against \gls{icl} \citep{Ghosh2026FineTuningICL}. \citet{Schulz2026UnravelingSyntax} decompose the language-modeling loss over a \gls{cfg}'s subgrammars, showing, analytically and empirically, that, throughout training, models acquire those substructures in parallel, in contrast to human children, who learn simpler subgrammars before complex ones.

\paragraph{Dyck languages.} Those are the languages of strings of $k$ bracket types that are correctly balanced and nested, a canonical object of language theory and a standard probe for unbounded recursion capabilities. \citet{Hahn2020TheoreticalLimitations} shows that under finite precision and Lipschitz-continuous attention a transformer decoder cannot recognize Dyck-2 on long-enough inputs. Our work does not contradict that finding, because we work on length-limited sentences, while Hahn's impossibility is asymptotic, showing that a fixed model cannot recognize the whole of an unbounded-length Dyck-2 language.

\paragraph{Scaling and emergence.}
When and how the abilities of \glspl{llm} appear as scale grows is an open question. One school defends emergent capabilities: absent in smaller models and appearing abruptly past a critical scale, thus resisting extrapolation from the smooth scaling laws of smaller models \citep{Wei2022EmergentAbilities}. A skeptical counter-current argues that much of the apparent suddenness is an artifact of discontinuous or nonlinear metrics (\eg, exact-match accuracy) and dissolves under continuous metrics (\eg, cross-entropy) \citep{Schaeffer2023Mirage}. A reconciling view holds that smooth scaling in pre-training loss can coexist with threshold-like jumps in downstream accuracy \citep{Chen2024DownstreamScaling}.

\section{Conclusion}
\label{sec:conclusion}
Our central finding is that step-by-step reasoning is not exclusive to frontier models: tiny transformers, orders of magnitude below the scale at which natural-language \gls{cot} becomes viable, already benefit from a primitive form of it. On two tasks built over the Dyck-$k$ languages, training on a protoreasoning trace substantially closes the gap between in- and out-of-distribution validity.

Because each model is tiny and each task carries two controllable axes of difficulty, we can grade generalization exactly and train thousands of models a day on a single compute node, a scale of experimentation that would be extremely cumbersome with larger frontier models.

Expressivity and generalization show very different results: our models reach perfect in-distribution validity even for over a hundred brackets and complex structure, yet cannot zero-shot generalize to out-of-distribution samples, particularly along the structural axis. The tasks are therefore easy to fit but hard to solve in a truly algorithmic way, making them appropriate to probe reasoning at our scales.

Out-of-distribution generalization is possible, particularly for interpolation, in which case the reasoning traces beat the vanilla non-reasoning format almost always, and often by a wide margin. Moreover, the gains the reasoning traces offer come from their content, not merely from the extra tokens they introduce.

Still, the extrapolation patterns show that a truly generalizable algorithmic solution remains elusive. Without implicit or explicit signals shaping the optimization, the model has no reason to prefer solutions more robust than those the in-distribution training loss and the model selection on out-of-distribution validation enforce. The absence of cross-transfer on mixed-task training further suggests that protoreasoning does not spontaneously become compositional or modular. Richer task families and trace formats that could stimulate such compositionality are a good focus for future work. We hope the framework itself --- cheap, fully controllable, and exactly gradable --- will prove as useful for the community's explorations as it was for the investigations of this paper.

\bibliography{bibliography/common,bibliography/surveys,bibliography/language_learning}
\bibliographystyle{tmlr}

\appendix
\section{Appendix}
\label{sec:appendix}
\subsection{Holdout patterns}
\label{sec:app-patterns}

\begin{figure}[ht]
  \centering
  \includegraphics[width=\textwidth]{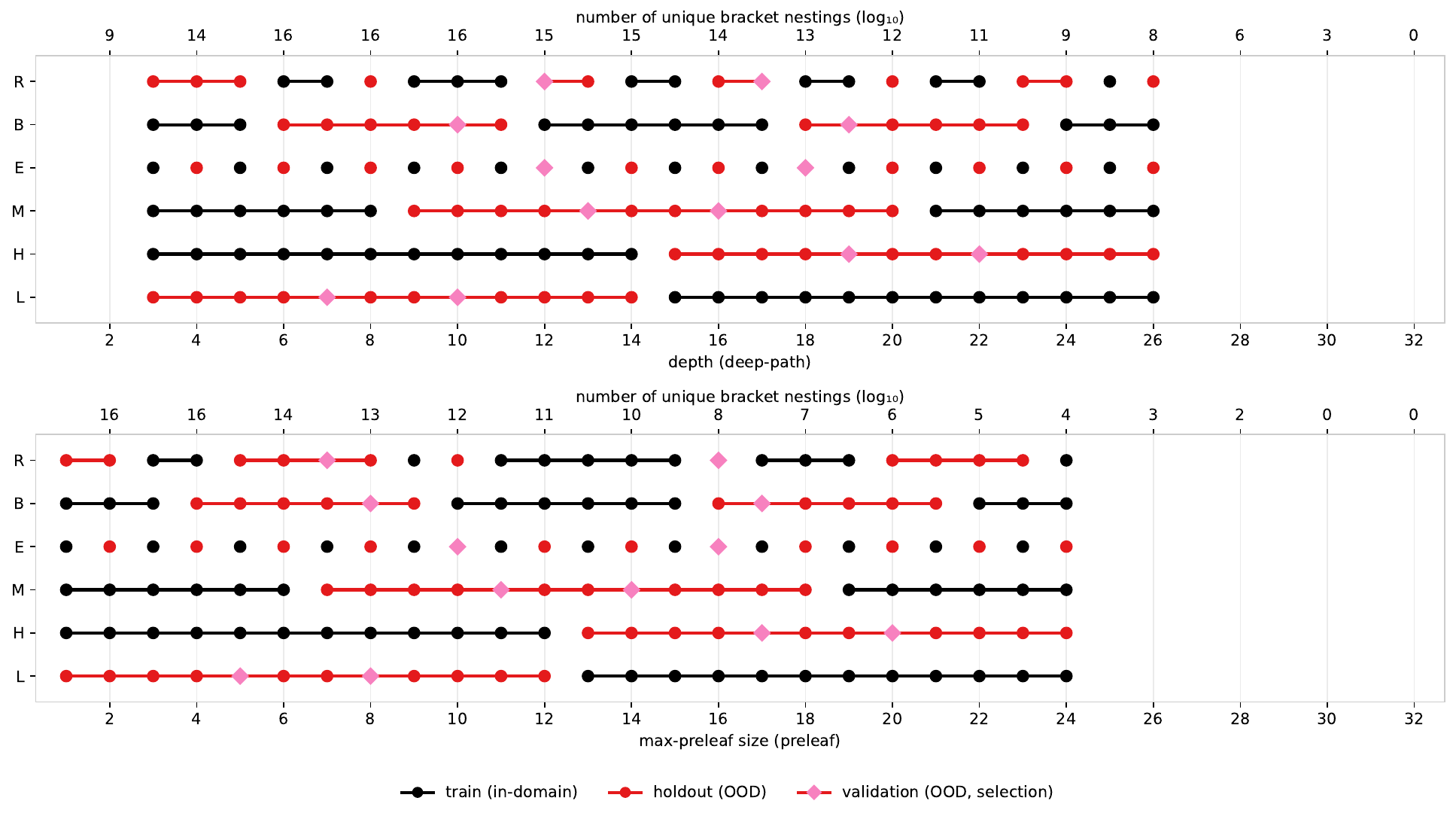}
  \caption{The six hold-out patterns (\textbf{R}andom, \textbf{B}locks, \textbf{E}ven, \textbf{M}iddle, \textbf{H}igh, and \textbf{L}ow) for single-task training (top: \glsentrylong{dp}/\glsentryshort{dp}; bottom: \glsentrylong{pl}/\glsentryshort{pl}), with each stratum (x-axis) colored as train (in-domain), holdout (OOD), or validation. \textbf{R}/\textbf{B}/\textbf{E}/\textbf{M} interleave seen and held-out strata (interpolation); \textbf{H}/\textbf{L} hold out extreme strata (extrapolation). The study covers depths and preleaf sizes with sufficient structural variation (log$_{10}$ count of unique bracket nestings on the top x-axes).}
  \label{fig:pattern-atlas}
\end{figure}

\Cref{fig:pattern-atlas} shows precisely how we split each task's strata into training, validation, and test sets for single-task training (\Cref{sec:res-patterns}). The top x-axis shows the $\log_{10}$ count of unique bracket nestings in each stratum. We select exactly 24 strata for each task, covering the region with sufficient complexity. Note that the actual number of sentences in each stratum is much larger because we have $k=4$ options for each bracket.

Mixed-task training (\Cref{sec:res-mixed}) follows patterns inspired by the above, but selects them among joint strata defined by both criteria (depth and maximum preleaf). Only strata with sufficient structural diversity (at least $10^4$ different bracket nestings) are considered, giving the selected region its triangular shape.

\begin{figure}[ht]
  \centering
  \includegraphics[width=\textwidth]{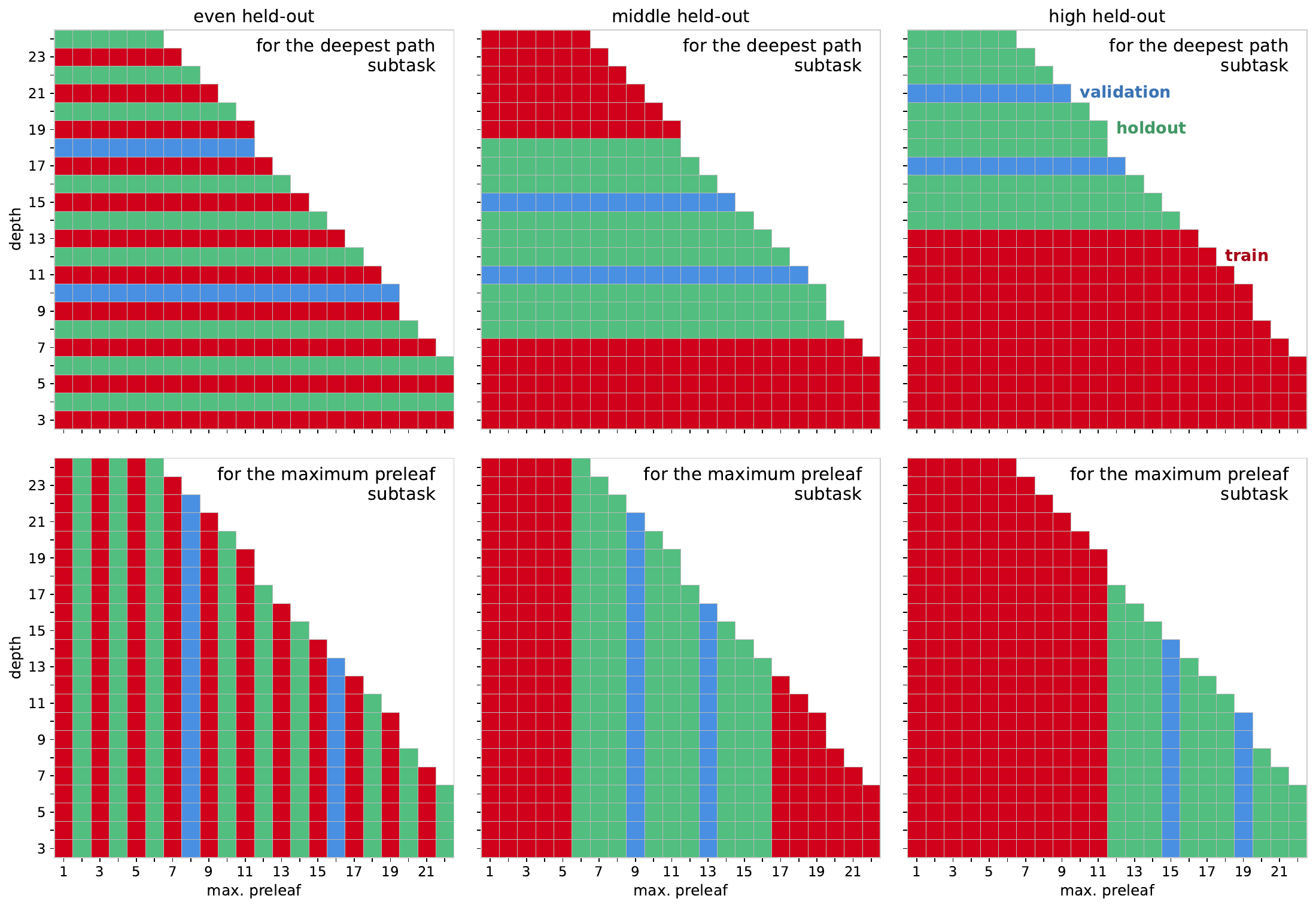}
  \caption{The three hold-out patterns (\textbf{E}ven-, \textbf{M}iddle-, and \textbf{H}igh-\textbf{M}ixed) for training by mixing both tasks. Each square is a joint stratum on both criteria (depth and maximum preleaf size). For a given pattern (columns), each row shows how the examples of a task (rows) are sampled. The model ends up exposed to all strata, but each task sees only its corresponding training strata.}
  \label{fig:mixed-partition-phasemap}
\end{figure}

\subsection{Reasoning-trace validity}
\label{sec:app-trace-validity}

\Cref{tab:trace-metrics} shows the detailed validity results for the trace format across the two tasks and six patterns. We track validity considering: output-only (output is valid, but trace is invalid), trace-only (trace is valid, but output is invalid), both (output and trace valid), and neither (output and trace invalid). In general, trace validity follows output validity, although some samples have correct output even with an invalid trace. The converse (valid trace with invalid output) is much rarer.

\begin{table}[ht]
  \centering
  \small
  \caption{Output and trace metrics for trace-format experiments on the six single-task patterns of \Cref{fig:pattern-atlas}. Columns show the fraction of valid generations that are valid only on Output, only on Trace, on both parts, or on neither. Discounting numerical rounding, the four metrics sum to 100\%. For the large majority of samples, output and trace validity match, although some samples have correct output despite an invalid trace. The converse is much rarer.}
  \label{tab:trace-metrics}
  \begin{btvtable}{C V C C C C V C C C C}
 & \SetCell[c=4]{c} \textsc{Deepest Path} &  &  &  & \SetCell[c=4]{c} \textsc{Max. Preleaf} &  &  &  \\
\textsc{Pattern} & \textsc{Output} & \textsc{Trace} & \textsc{Both} & \textsc{Neither} & \textsc{Output} & \textsc{Trace} & \textsc{Both} & \textsc{Neither} \\
\btvheadrule
Random & $5.7 \pm 1.1$  & $0.1 \pm 0.1$ & $89.7 \pm 2.7$ & $4.6 \pm 2.1$  & $2.3 \pm 0.3$  & $0.3 \pm 0.5$  & $95.5 \pm 0.6$ & $1.9 \pm 0.4$  \\
Blocks & $6.0 \pm 1.4$  & $0.0 \pm 0.0$ & $92.6 \pm 1.7$ & $1.3 \pm 0.4$  & $2.7 \pm 0.9$  & $0.1 \pm 0.1$  & $94.6 \pm 1.0$ & $2.6 \pm 0.5$  \\
Even   & $4.1 \pm 2.1$  & $0.0 \pm 0.0$ & $94.7 \pm 2.9$ & $1.2 \pm 0.9$  & $3.5 \pm 0.7$  & $0.1 \pm 0.1$  & $94.1 \pm 1.1$ & $2.2 \pm 0.5$  \\
Middle & $6.7 \pm 6.8$  & $0.2 \pm 0.2$ & $91.3 \pm 8.8$ & $1.9 \pm 2.0$  & $4.4 \pm 0.7$  & $1.4 \pm 0.8$  & $92.7 \pm 1.1$ & $1.5 \pm 0.5$  \\
High   & $12.4 \pm 4.4$ & $0.2 \pm 0.2$ & $54.3 \pm 8.8$ & $33.2 \pm 9.4$ & $5.1 \pm 5.4$  & $17.0 \pm 7.2$ & $57.1 \pm 9.4$ & $20.8 \pm 8.4$ \\
Low    & $13.0 \pm 4.1$ & $0.3 \pm 0.4$ & $8.6 \pm 7.8$  & $78.1 \pm 7.6$ & $27.4 \pm 4.8$ & $1.3 \pm 1.2$  & $13.6 \pm 7.9$ & $57.7 \pm 6.3$ \\
\end{btvtable}

\end{table}

\subsection{Positional augmentation and step dropout}
\label{sec:app-positional-augment}

\paragraph{SkipAlign.}
To train length generalization without training on long sequences, we apply SkipAlign~\citep{Wu2024SkipAlign}, a refinement of PoSE~\citep{Zhu2024PoSE} that remaps the position indices of a short sequence so that they span a much larger range.
We insert a positional gap before the token that \emph{starts} each reasoning step --- the token immediately after a newline inside the \texttt{<BOT>}$\dots$\texttt{<EOT>} region, plus one gap at the start of reasoning --- so the gaps land on semantic step boundaries rather than arbitrary positions. For vanilla samples, there is a single gap opportunity, between the task separator and the output.

Our SkipAlign implementation samples a target length from a log-uniform distribution, ranging from the natural length of the sample (without gaps) to the maximum possible length for a sample across all strata of that task and format. We then sample how to split the gap budget across all boundaries from a sparse Dirichlet ($\alpha{=}0.4$).
The augmentation is a pure remapping of position indices: the tokens themselves are unchanged. It encourages the model to consider positions spanning the entire range of possible input lengths, without requiring overly long contexts.

\paragraph{Step dropout.}
For trace models, step dropout independently drops each reasoning step with probability $p=0.15$; it only deletes steps, never reorders or invents them, and leaves the final output unchanged. This forces the model to recover the answer from an arbitrary subsequence of the reasoning steps instead of memorizing a fixed-length scaffold.
We score trace validity accordingly: a generated trace is valid iff its steps form an order-preserving subsequence of the reference trace's steps (each retained step is still matched token-for-token).

\subsection{Implementation Details}
 \label{sec:app-details}

\paragraph{Model.}
All experiments use a Llama 2-like decoder-only transformer with rotary positional embeddings, multi-head attention, grouped-query attention disabled, and RMSNorm. Except for \Cref{sec:res-size-scaling}, all experiments set width $w{=}128$, depth $L{=}4$, and 16 query/key/value heads, resulting in $\sim$1M parameters, orders of magnitude below even first-generation transformers.

\paragraph{Hyperparameters.}
Each treatment's AdamW parameters (learning rate and weight decay) come from an 8-trial quasi-random (Sobol) search selected according to output validity on the validation strata, which are disjoint from both training and test (\Cref{sec:app-patterns}). The expressivity studies, which train on a single stratum, select hyperparameters in-domain. The search space for the parameters has learning rate from $10^{-4}$ to $10^{-2}$ and weight decay from $10^{-7}$ to $10^{-1}$. Fixed parameters are AdamW's $\beta_1=0.9$, $\beta_2=0.95$, and $\epsilon=10^{-6.5}$.

\paragraph{Training length and schedule.}
We train all models with a batch size of 256 for 6K steps, with a warmup--stable--decay learning-rate schedule (600 warmup and 600 decay steps).
We pretrain each model for 2K steps, during which the autoregressive loss applies to the entire sample, and then fine-tune it for 4K steps, with the loss accounting only for the generated part (including the trace, if present). We apply SkipAlign for 1.3K steps at the beginning of the fine-tuning phase.

\paragraph{Evaluation.}
Unless noted otherwise, we report greedy-decoding exact-match accuracy at the best-validation checkpoint. Evaluation is over the holdout strata (out of domain) for the generalization experiments and over the training strata (in domain) for the expressivity experiments.
We score output validity on the validation strata every 500 training steps and evaluate the checkpoint that maximizes it, breaking ties toward the earliest step; the same best-checkpoint criterion scores each trial of the hyperparameter search. For trace models, selection always ranks on the output-only validity.
All validation is over a fixed set of 128 prompts per stratum.

\paragraph{Compute.}
The full study comprises 186 treatments trained over 10 seeds (1860 final training runs plus 1488 hyperparameter-search training runs). The complete study takes a little over a day of wall-clock time on a single multi-GPU node with $7\times$ NVIDIA B200 GPUs.

\subsection{Model and language scaling}
\label{sec:res-size-scaling}

In this section, we measure the ability of different small models to express and generalize both tasks on languages of different lengths, all for the vanilla data format. For each language, we split the strata randomly as training, validation, and test (the Random pattern of \Cref{sec:app-patterns}). As usual, we measure expressivity on different in-domain sentences sampled from the training strata and generalization on out-of-domain sentences sampled from the test strata.

\begin{figure}[ht]
  \centering
  \scalebox{0.9}[0.8]{\includegraphics[width=\textwidth]{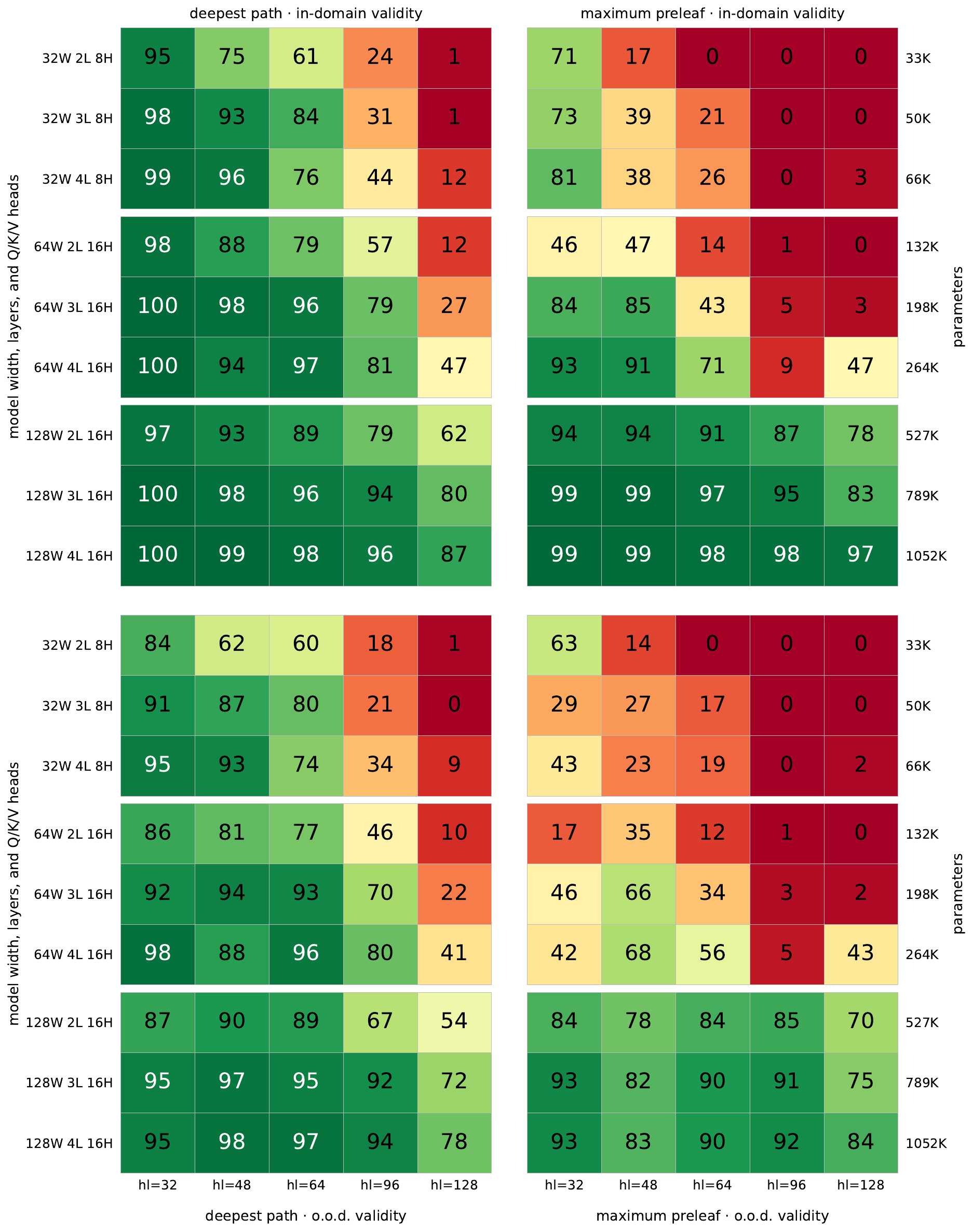}}
  \caption{In-domain expressivity (top row) and out-of-domain generalization for models trained and tested on randomly split structural strata. Model configuration appears on the left axis (width, number of layers, number of Q/K/V heads), parameter count on the right axis. The bottom axis shows the half length of the language parameters used to train and evaluate the model.}
  \label{fig:model-sweep-phasemap}
\end{figure}

\Cref{fig:model-sweep-phasemap} shows that, among very small models, there is a relatively sharp frontier between those that can express the task for a certain length and those that cannot. It also shows that generalization is at least as hard as expressivity. For the bulk of this work, we used half-length~= 32 and a model with residual width~= 128, 4 attention layers, and 16 Q/K/V attention heads, a scenario where expressivity is near-perfect but generalization already poses some challenge, even for the relatively easy Random pattern.

\end{document}